# Malaria Detection from Blood Cell Images Using XceptionNet

Warisa Nusrat, Mostafijur Rahman[†], Ayatullah Faruk Mollah

*Department of Computer Science and Engineering, Aliah University,*
*IIA/27 Newtown, Kolkata 700160, India*

[†] Corresponding author: mostafijurrcse@gmail.com

**Abstract**. Malaria, which primarily spreads with the bite of female anopheles mosquitos, often leads to death of people - specifically children in the age-group of 0-5 years. Clinical experts identify malaria by observing RBCs in blood smeared images with a microscope. Lack of adequate professional knowledge and skills, and most importantly manual involvement may cause incorrect diagnosis. Therefore, computer aided automatic diagnosis stands as a preferred substitute. In this paper, well-demonstrated deep networks have been applied to extract deep intrinsic features from blood cell images and thereafter classify them as malaria infected or healthy cells. Among the six deep convolutional networks employed in this work viz. AlexNet, XceptionNet, VGG-19, Residual Attention Network, DenseNet-121 and Custom-CNN. Residual Attention Network and XceptionNet perform relatively better than the rest on a publicly available malaria cell image dataset. They yield an average accuracy of 97.28% and 97.55% respectively, that surpasses other related methods on the same dataset. These findings highly encourage the reality of deep learning driven method for automatic and reliable detection of malaria while minimizing direct manual involvement.

## 1. Introduction

In recent years, malaria has been one of the most dangerous diseases in the world. Mainly malaria is spread by the bite of female Anopheles mosquitoes. After the bite, protozoan parasites are spread in the red blood cells of humans, then this parasite continues its reproduction cycle. A total of five types of Plasmodium species are found, among them the Plasmodium vivax species is the main reason to spread malaria. A lot of people, mostly children between the ages of 0 to 5 years, have died of this dangerous disease. Traditional microscope-based malaria diagnosis of blood smear, this process needs highly experienced workers to detect malaria from blood cells. This process is prone to errors, especially when there is a lack of resources with inexperienced workers with outdated equipment and poor maintenance, which might be the cause of false positives and false negatives testing results, which can affect patient treatment, and this process takes a long time.

Focusing on modern approaches, which reduce human involvement for malaria diagnosis and improve automated analysis, advances technology, enhances accuracy and reliability. To overcome all these challenges, we used advanced deep convolutional networks, which have possibilities for automated detection of malaria parasites from the Blood Cell Images dataset. Our proposed approach, XceptionNet, which is pre-trained and customized additional layers with depth wise separable convolutional, provides high efficiency, accuracy, reliability and reduces the dependency on manual intervention and provides automated and scalable solutions to enhance global malaria diagnosis and treatment.

Additionally, five other deep convolutional networks have been employed viz. DenseNet-121, VGG-19, AlexNet, Custom-CNN and Residual Attention Network to substantiate the superiority of XceptionNet in reliable and accurate prediction for malaria detection. The paper is organized as follows: Section 2 presents an overview of related works along with motivation for the current work. Deep networks have been described in Section 3. Experiments, results and analysis have been shown in Section 4 while conclusion is drawn and further works have been outlined in Section 5.

## 2. Overview of Related Works

The use of deep learning CNN models for malaria detection has been studied, researchers are focusing on improving accuracy in diagnosing malaria from blood smear images. Rajaraman et al. [1] applied a pre-trained CNN model on thin blood smear images to detect malaria. The aim of this approach is enhanced diagnostic accuracy and more reliable identification of malaria disease. Rahman et al. [2] explore three CNN architectures, AlexNet, ResNet-50 and VGG19 for automated parasite detection on diverse blood smear images. The main advantage of this approach is that it applies an advanced CNN-based model over traditional methods, which are labor-intensive and time-consuming. Singha et al. [3] highlighted the potential of deep learning models in malaria diagnosis. Deep CNN-based approaches applied to labelled blood smear images improve the testing accuracy, specificity and sensitivity. Wang et al. [4] used a Residual Attention Network with deep architecture for malaria parasite detection, this approach has an attention mechanism with a residual learning framework, which improves classification results. Desai et al. [5] proposed a hybrid approach with a combination of deep learning models with machine learning classifiers on a context-based image dataset, this approach bridges the gap between feature extraction and classification tasks.

Some challenges are identified in applying deep learning for parasite detection, requiring high computational demand for complex architecture like attention-based networks, which need highly constrained resource settings, and some models used limited-size datasets and tried to achieve higher performance, that might be prone to overfitting. In this paper, our proposed approach is used, depth wise separable convolutions in the XceptionNet model, to ensure computational efficiency and that the architecture is suitable for low-resource environments. This architecture has integrated capabilities to extract features and classify the dataset. The layer of this architecture can more perfectly extract features and is suitable for multiclass classification. This approach is a scalable solution to global health challenges and significantly improves diagnostic accuracy, especially in resource-poor areas.

## 3. Architecture Design

In this paper, different types of CNN architectures are used to classify parasitized and uninfected blood cells. The architectures are AlexNet, XceptionNet, VGG-19, Residual Attention Network, DenseNet-121 and Custom-CNN. In these architectures, one or more convolutional layer, max pooling layer, average pooling layer, global average pooling, flatten layer, dense layer, ReLU activation function, softmax activation function, sigmoid activation function and loss function called 'categorical-cross entropy', 'Adam' optimizer, etc. were used.

### 3.1 DenseNet-121

The DenseNet-121 architecture is used for feature extraction and classification tasks. This architecture configuration is convolutional-based, which is efficient for feature extraction. As shown in Fig. 1, the input size shape 128x128x3. In this architecture, an executed, fully connected layer ensures that the pre-trained weight remains unchanged during the training period. This approach is able to learn the feature representation of classification tasks. A custom layer is added to provide adaptability. The first global average pooling 2D layer is used to reduce the spatial dimensions. In the features map, this layer ensures that a compact representation of extracted features is created. The next fully connected layer with a ReLU activation function contains 512 neurons, which provides the ability to learn the complex patterns of the classification task. The dense layer is used for regularization. A dropout layer with a 0.5 rate is used to prevent the overfitting during training by randomly deactivating 50% of the neurons. Finally, the output layer with sigmoid activation contains 2 neurons, which provides the probability of two classes. This architecture-built capability to model for robust feature extraction and reduce the computational cost by freezing the DenseNet base. While retaining the generation of pertained features.

The dense layer has 512 trainable parameters, and the output layer has a series of additional parameters. This architecture is suitable for small and medium-sized datasets.

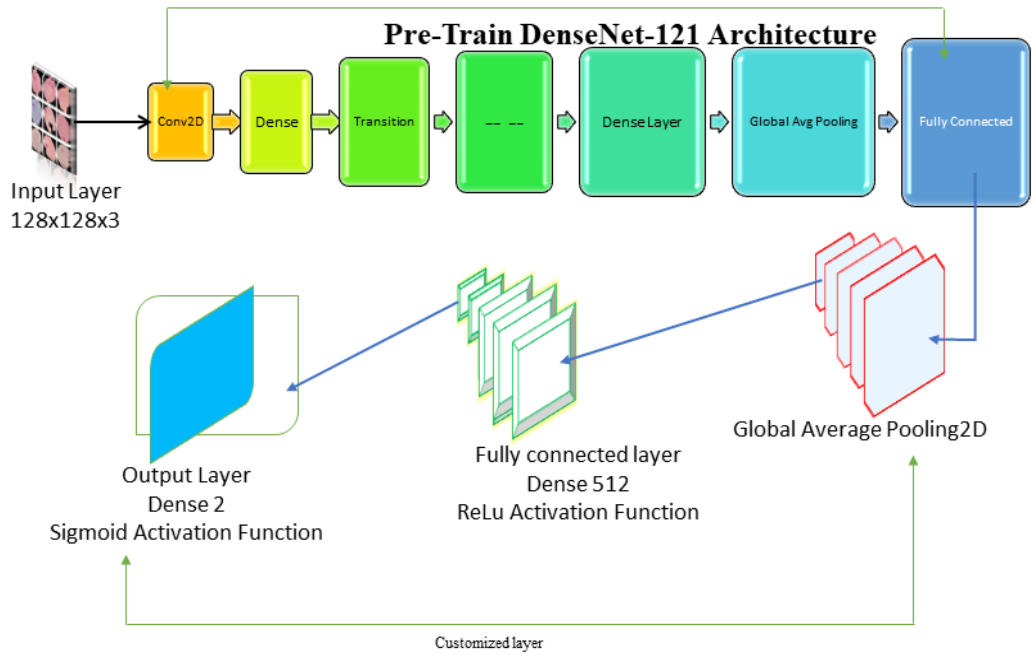

**Fig. 1** Block diagram of DenseNet-121

## 3.2 VGG-19

The VGG19 architecture combines a pre-trained VGG19 base with a custom layer for classification tasks. This architecture is an extended version of the VGG16 model. This architecture has 19 layers, including 16 convolution layers and 3 fully connected layers. The convolution layer has 3x3 filters with a stride of 1 and padding used for spatial resolution. This architecture used convolutional layers with 20 million parameters. These are non-trainable convolutional layers used for trained attributes, where pretrained weights remain unchanged. VGG19 with a 4D tensor shape of 4x4x5124 filters. This model has 24.2 million parameters; among them, 4.2 million are trainable with the corresponding custom dense layer. The flatten layer with the ReLU activation function contains 512 neurons. The final dense layer with a sigmoid activation function contains 2 neurons, this layer is used for binary classification. This architecture's efficient design is capable of the model for feature extraction, this design also reduced the computational burden for the classification task. This architecture has nearly 143 million parameters; that's why this model is computationally expensive. This model is resource-intensive compared to other modern architectures, it provides robust representation. This model is also used for transfer learning and object detection.

## 3.3 AlexNet

AlexNet architecture with modern image dimensions for classification tasks and feature extraction. This architecture is also balanced in the classification performance and features extraction. This architecture is used for spatial feature extraction on input images of size 128×128×3. Firstly, convolution layers are applied with 96 filters with a kernel size of 11x11 and a stride 4. Max pooling layer of 3x3 size filter with a stride of 2 for reducing dimensions. Second convolution layers with filter size 256 and with kernel size 5x5 with the same padding used for maintaining spatial dimensions. Then three convolutional layers are used: the first of two convolutional layers with 384 filter sizes and the last convolutional layer has 256 filters with a kernel size of 3x3 and the same padding. This architecture design is a little bit complex with fine-grained features with a block size of 3x3. Fully connected layers are used for high-dimensional representation for classification tasks. Dense layer containing 4096 neurons with a ReLU activation function. A dropout layer is used to prevent overfitting during training by randomly deactivating neurons. Dense layer with the same filter size and the same activation function for learning representation. The dropout layer is used for regularization. The final dense layer with a sigmoid activation function contains 2 neurons for the probability output. This model architecture used a total 60 million parameters, with the convolutional layers with large-size kernels and a series of filters. A fully connected layer with two dense layers added 4096 neurons and millions of additional parameters, used to ensure that the model perfectly learns the complex relationship between data.

### 3.4 Custom-CNN

A custom-CNN architecture is built with a convolutional neural network by using Keras (sequential) and API. This architecture is designed for classification of input shape 128×128×3, where it represents height and weight for input images, and 3 denotes RGB color channels. This architecture has three convolutional blocks for feature extraction. Each block follows the batch normalization. These blocks increase filter size, like (32, 64, and 128), and kernel size also. Batch normalization normalized the activation and improved stability. Max pooling layer is used for feature mapping and reducing the spatial dimensions. A dropout layer is used to prevent overfitting. After feature extraction, the output of the final convolutional block is transferred into a 1D vector by the flatten layer. The first dense layer has 128 neurons with the ReLU activation function. The final dense layer contains 2 neurons with a sigmoid activation function that are used to predict the probability of two classes, this layer is provided a higher probability for each class. This model architecture is designed for balancing complexity and performance. The first convolutional block has 1024 parameters, the second block has 18752 parameters, and the third block has 74368 parameters. The fully connected layer has 258 parameters. This architecture is compatible with medium-sized datasets and captures the hierarchical features. This architecture is suitable for binary classification.

### 3.5 Residual Attention Network

In this paper, a residual attention network is used to detect the malaria-infected blood cell image. In this architecture, it has advanced attention mechanisms, and as a result, it optimizes in the training period. Fig. 2 showing the Residual Attention Network model working flow, at first the input layer shape is 128x128x3, then the three convolutional layers were added with conv2D, 64 filters, 3x3, batch normalization, and ReLU activation function. After that, the global average pooling layer is passed through a fully connected layer with 32 units and ReLU activation function. After that, added the dense layer, which is a fully connected layer with 64 units and a sigmoid activation function. The values of the sigmoid activation function are between 0 and 1.

$$\text{sigmoid}(x) = \frac{1}{1+e^{-x}}$$

Residual Attention Network architecture is very complex, but it still works well on certain datasets. The Residual Attention Network model has a risk of overfitting. The Residual Attention Network model integrates residual connection and attention mechanism. This model has the main motive to improve network ability to focus on enhanced information. Residual connections allow smooth flow during training. Attention mechanisms allow the network to focus on crucial parts of input data, enhancing its ability to differentiate features and focus on capturing more relevant information. This attention mechanism allows for the creation of a hierarchical representation of features that captures low-level to high-level semantic information. Residual Attention Network architecture mechanisms improve overall performance, but this model also has limitations, like attention mechanisms and residual connections, which take a lot of time for training and are computationally expensive.

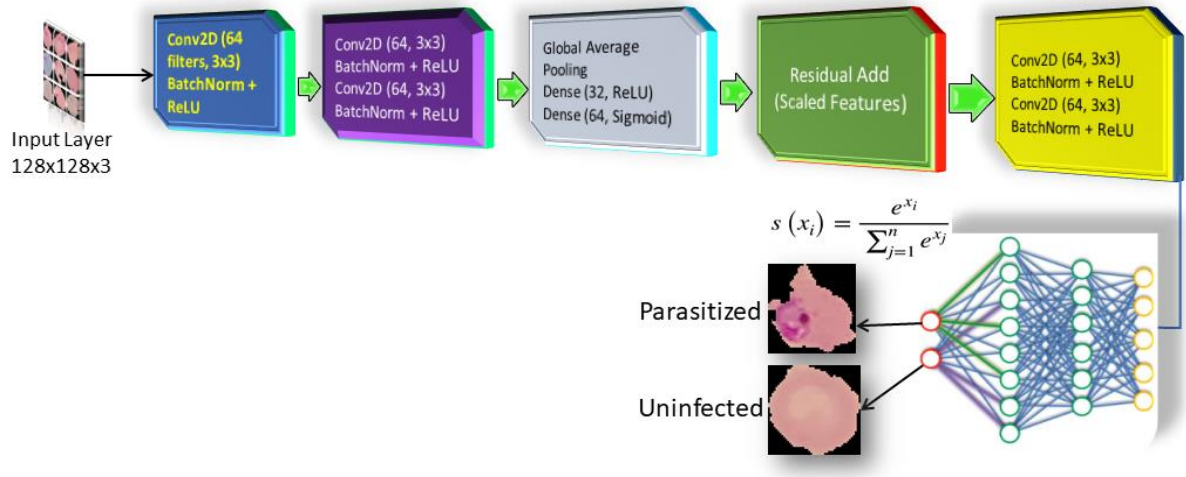

**Fig. 2** Block diagram of Residual Attention Network

### 3.6 XceptionNet

The XceptionNet architecture is used for detecting the malaria-infected cell. This architecture is well suited for medical image classification as well as malaria blood cell image detection. It has integrated capabilities to extract features and classify the dataset. It is mainly used to train large scale datasets. The efficiency and performance of the architecture depend on depth-wise convolution layers. These layers are able to extract high-level features. Fig. 3 showing the XceptionNet architecture block diagram, this architecture used pre-trained and customized additional layer design in this work. The input layer dimension is 128x128x3 adapted to the malaria datasets. The features extraction base model is loaded with the include_top= False, which means it allows the use of the feature's extractor, which excludes the fully connected layers for feature extraction. Global max pooling layers are used for feature maps and reduce the dimensions of 1D vectors.

After that, adding the custom layer. First of all, a flattening layer is used to convert 1D feature vectors to arrays. Then a dropout layer 0.3 means randomly disabling 30% of neurons during training, and it's used to prevent overfitting the model. The first dense layer, which means a fully connected layer with 128 units of neurons and ReLU activation functions, is used to learn the complex patterns. Batch normalization is used to provide stability and acceleration to training methods and dropout layer 0.3. The second dense layer with 64 units of neurons and ReLU activation functions is used to extract features more perfectly, and the last dropout layer is 0.25. A final dense layer with 2 units of neurons (for binary classification of malaria infected or uninfected) and a softmax activation function is used for the output probability of classes. Categorical cross entropy is a loss function that is suitable for multiclass classification problems. Adam optimizer is used for efficient and stable convergence.

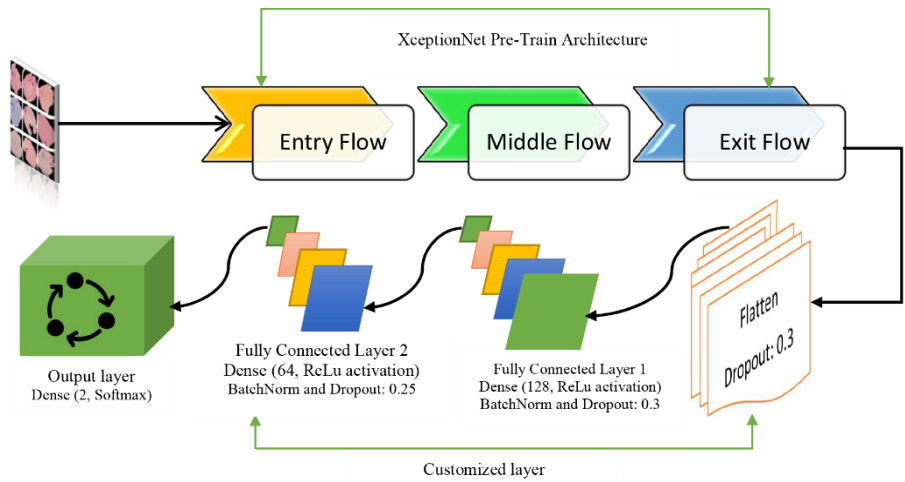

**Fig. 3** Block diagram of XceptionNet Architecture

## 4. Experimental Results and Analysis

The methods discussed in Section 3 were implemented on an NVIDIA Tesla T4 GPU with 8 GB RAM using Python libraries, such as pandas, numpy, matplotlib, cv2, TensorFlow and Keras. The learning rate was set to 0.001 with batch size of 32 and Adam optimizer. Evaluation metrics included accuracy, precision, recall, f1-score, ROC-AUC and RMSE.

### 4.1 Dataset Outline

The malaria blood cell images dataset [7], publicly available in Kaggle, is considered for the experiments carried out in this work. It is released by the U.S. National Library of Medicine, with the main objective of fostering the development of an automated malaria detection system. This dataset comprises of 13,780 parasitized (infected with the malaria parasite) and 13,778 uninfected (normal healthy cell) cell images. All these 27,558 images are of 128x128 pixels in PNG format. This dataset is split into training, validation and test sets in the ratio of 8:1:1. Thus, the number of training samples is 22,046, that of validation and testing are 2,756 each. Fig. 4 showing the malaria blood cell parasite and uninfected

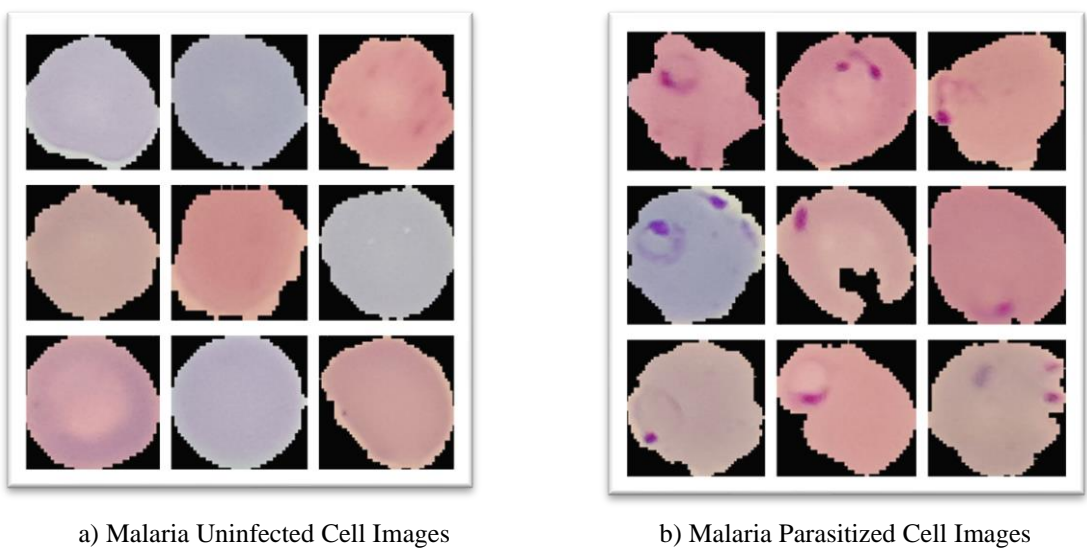

a) Malaria Uninfected Cell Images          b) Malaria Parasitized Cell Images

**Fig. 4** Samples of malaria cell image from Malaria Blood Cell Images [7].

## 4.2 Model Performance and Analysis

Training and testing accuracy and loss comparison of different architectures. DenseNet-121 has lower accuracy and higher loss compared to all architectures in this paper. This model is not well suited for malaria blood cell images. This model the minimal gap between train and test matrices. The training accuracy is 0.9088, and the training loss is 0.2676, the testing accuracy is 0.8976, and the testing loss is 0.2676. Also, VGG-19 has lower performance compared to other advanced architectures of this paper. The training accuracy is 0.9263, and the training loss is 0.1882, and the testing accuracy is 0.9048, and the testing loss is 0.2685. The AlexNet model does not perform well, with a training accuracy of 0.9687 and training loss of 0.1033 and testing accuracy of 0.9592 and testing loss of 0.1787. Custom CNN has slightly better performance. This model has a training accuracy of 0.9789 and a training loss of 0.0610 and a testing accuracy of 0.9611 and a testing loss of 0.1238.

The Residual Attention Network model performance is a little bit lower than XceptionNet, this model has a slightly gap between train and test matrices, which shows the strong generalization capability. This model has a training accuracy of 0.9866 and a training loss of 0.0657 and a testing accuracy of 0.9728 and a testing loss of 0.0886, and the XceptionNet model has the best performance; it has a minimal gap between training and testing accuracy, which indicates that there is negligible overfitting and robust generalization, which is well suited for malaria infected cells detection. This model has a training accuracy of 0.9850 and a training loss of 0.0506 and a testing accuracy of 0.9755 and a testing loss of 0.800.

**Table 1.** Malaria parasite detection performance for various leading deep networks.

| Method | Accuracy | Precision | Recall | F1-Score | AUC-ROC | RMSE |
|---|---|---|---|---|---|---|
| DenseNet-121 | 89.76% | 0.8988 | 0.8976 | 0.8976 | 0.8980 | 0.3200 |
| VGG-19 | 90.48% | 0.9056 | 0.9048 | 0.9048 | 0.9050 | 0.3085 |
| AlexNet | 95.92% | 0.9592 | 0.9591 | 0.9593 | 0.9592 | 0.2019 |
| CNN [Custom] | 96.11% | 0.9610 | 0.9611 | 0.9610 | 0.9610 | 0.1816 |
| Res_Attention_Net | 97.28% | 0.9705 | 0.9716 | 0.9733 | 0.9728 | 0.1649 |
| XceptionNet | **97.55%** | 0.9755 | 0.9755 | 0.9755 | 0.9756 | 0.1564 |

This work explored various deep convolutional network architectures to accurately detect malaria-infected cells and evaluated their performance metrics. According to Table 1, it is observed DenseNet-121 demonstrated the lowest test accuracy, achieving 89.76%. This is relatively inefficient performance as compared to other architectures used in this paper. VGG-19, AlexNet and Custom-CNN with their test accuracies of 90.48%, 95.92% and 96.11% respectively. However, custom-CNN architecture performance is improved. The XceptionNet architecture achieved the highest accuracy 97.55% and high consistency across precision, recall, and f1-score. Residual attention networks also have good accuracy, 97.28%. XceptionNet and residual attention network architecture performance are the best for malaria detection, these architectures provided reliable and accurate predictions for malaria detection.

Table 2. Accuracy and loss comparison for different models.

| Models | Training Accuracy | | Testing Accuracy | |
|---|---|---|---|---|
| | Accuracy | Loss | Accuracy | Loss |
| DenseNet-121 | 0.9088 | 0.2676 | 0.8976 | 0.2997 |
| VGG-19 | 0.9263 | 0.1882 | 0.9048 | 0.2685 |
| AlexNet | 0.9687 | 0.1033 | 0.9592 | 0.1787 |
| CNN [Custom] | 0.9789 | 0.0610 | 0.9611 | 0.1238 |
| Res_Attention_Net | 0.9866 | 0.0657 | 0.9728 | 0.0886 |
| XceptionNet | 0.9850 | 0.0506 | 0.9755 | 0.0800 |

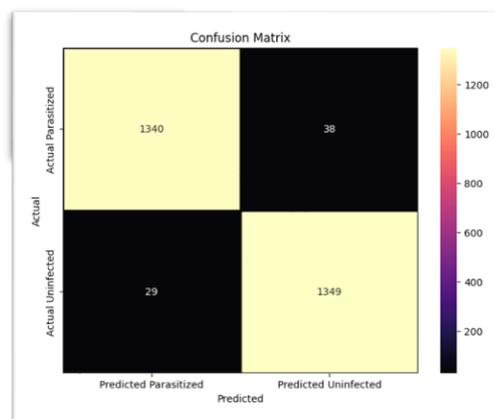

**Fig. 5** Confusion matrix of XceptionNet architecture.

A confusion matrix is used to evaluate the performance metrics of a classification architecture by comparing its predicted results and actual outcomes showing in Fig. 5.

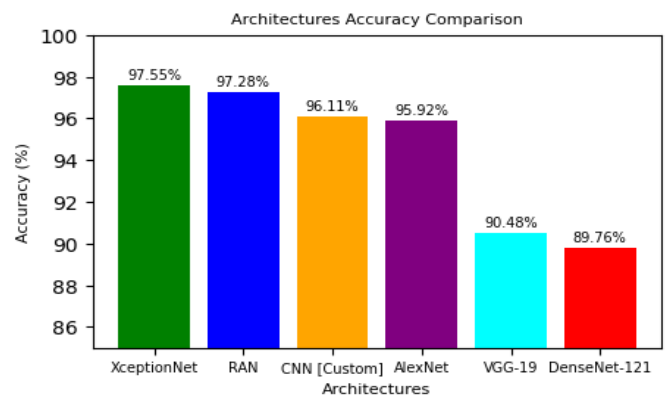

**Fig. 6** Accuracy comparison of employed architectures.

According to Fig. 6, this histogram is a visualization of the accuracy of various deep learning architectures for detecting malaria-infected cells. The architectures in this paper are used for comparison, with their corresponding accuracy displayed on the Y-axis. Each bar represents an architecture, and its height reflects the accuracy achieved. Here the highest accuracy is achieved by the proposed architecture.

### 4.3 Comparative Study and Discussion

The comprehensive analysis of the state-of-the-art methods shows the performance improvement by our proposed approach. Our approach has been improved by following some of these methods. Like AlexNet [1], which achieved 93.70% accuracy with a corresponding f1-score of 0.9370, MCC of 0.8720, and AUC of 0.9810. The next method is CNNEx-SVM [3], which showed test accuracy of 94.77% with precision of 0.9213, recall of 0.9515, an F1 score of 0.9501, and an MCC of 0.8925. Another method is ResNet-50 [10], which tests performance achieved 95.70% with a precision of 0.9690, recall of 0.9450, F1 score of 0.9570, and AUC of 0.9900. K-fold cross-validation [13], which has 89.10%.

**Table 3.** Performance comparison with the state-of-the-art methods

| Method | ACC | Precision | Recall | F1- SC | MCC | AUC | Sensitivity | Specificity |
|---|---|---|---|---|---|---|---|---|
| Alex Net [1] | 93.70 | - | - | 0.9370 | 0.8720 | 0.9810 | 0.940 | 0.9330 |
| CNNEx-SVM [3] | 94.77 | 0.9213 | 0.9515 | 0.9501 | 0.8925 | 0.9101 | - | - |
| ResNet-50 [10] | 95.70 | 0.9690 | 0.9450 | 0.9570 | 0.9120 | 0.9900 | - | - |
| AlexNet_FC7[11] | 96.18 | - | - | - | - | - | - | - |
| Normalized cross_ correlation [12] | 95.00 | - | - | - | - | - | - | - |
| K flod cross validation [13] | 89.10 | - | - | - | - | - | 0.9390 | 0.8310 |
| VGG-19 [14] | 93.89 | 0.9377 | - | 0.9333 | - | - | 0.9295 | 0.9469 |
| CNN [15] | 88.00 | 0.8800 | 0.8800 | 0.8800 | - | - | - | - |
| **Ours** | **97.55** | **0.9755** | **0.9755** | **0.9755** | - | **0.9756** | - | - |

All these approaches used advanced methods, but compared with our method has better performance. Our proposed approach has excellent performance, which achieves 97.55% accuracy, with precision, recall, and F1 score all at 0.9755. Our proposed approach significantly improved ResNet-50 [10] and AlexNet [1]. Our approach has an accuracy of 0.9756, which indicates reliability and robustness for the classification task. Another method of our approach is VGG19, which highlights strong sensitivity and specificity. Our proposed approach provides consistent and balanced performance, the results of our proposed approach established a new state of the art in the evaluation domain.

### 5. Conclusion

In this work, the power of deep convolutional networks is tested in detecting malaria parasite from blood cell images. Among all the networks considered, XceptionNet and Residual Attention Network produced quite impressive performance (97.55% and 97.28% respectively on test sets). All these architectures highlight the capability of generalization, minimal overfitting and maximum efficiency. This approach is a way-forward for the future of AI-driven diagnostic roles in global health challenges. However, it applies on well-segmented blood cell images. Given a raw blood-smeared image, segmentation must be done prior to classification with XceptionNet. In future, exploration on accurate segmentation is needed and XceptionNet may be applied on other blood smeared images to examine its performance.

# References

1. S. Rajaraman, S. K. Antani, M. Poostchi, K. Silamut, M. A. Hossain, R. J. Maude, S. Jaeger, and G. R. Thoma, "Pre-trained convolutional neural networks as feature extractors toward improved malaria parasite detection in thin blood smear images," PeerJ, vol. 6, p. e4568, 2018.
2. A. Rahman, H. Zunair, M. S. Rahman, J. Q. Yuki, S. Biswas, M. A. Alam, N. B. Alam, and M. R. C. Mahdy, "Improving malaria parasite detection from red blood cell using deep convolutional neural networks," arXiv preprint arXiv:1907.10418, 2019.
3. D. Sinha and M. El-Sharkawy, "Thin MobileNet: An enhanced MobileNet architecture," in 2019 IEEE 10th Annual Ubiquitous Computing, Electronics & Mobile Communication Conference (UEMCON), 2019, pp. 280–285.
4. F. Wang, M. Jiang, C. Qian, S. Yang, C. Li, H. Zhang, X. Wang, and X. Tang, "Residual attention network for image classification," in Proc. IEEE Conf. Comput. Vis. Pattern Recognit., 2017, pp. 3156–3164.
5. P. Desai, J. Pujari, C. Sujatha, A. Kamble, and A. Kambli, "Hybrid approach for content-based image retrieval using VGG16 layered architecture and SVM: An application of deep learning," SN Computer Science, vol. 2, no. 3, p. 170, 2021.
6. World Health Organization, World Malaria Report. [Online]. Available: https://www.who.int/publications/i/item/9789240015791, 2020
7. Malaria Cell Images Dataset. [Online]. Available: https://www.kaggle.com/datasets/iarunava/cell-images-for-detecting-malaria, Accessed: Nov. 2024.
8. M. S. N. Raju and B. S. Rao, "Colorectal multi-class image classification using deep learning models," Int. J. Electr. Comput. Eng. (IJECE), vol. 11, no. 1, pp. 195–200, Feb. 2022.
9. M. R. Islam and A. Matin, "Detection of COVID-19 from CT image by the novel LeNet-5 CNN architecture," in 2020 23rd Int. Conf. Comput. Inf. Technol. (ICCIT), 2020, pp. 1–5.
10. S. Rajaraman, S. K. Antani, M. Poostchi, K. Silamut, M. A. Hossain, R. J. Maude, S. Jaeger, and G. R. Thoma, "Pre-trained convolutional neural networks as feature extractors toward improved malaria parasite detection in thin blood smear images," PeerJ, vol. 6, p. e4568, 2018.
11. W. Kudisthalert, K. Pasupa, and S. Tongsima, "Counting and classification of malarial parasite from giemsa-stained thin film images," IEEE Access, vol. 8, pp. 78663–78682, 2020.
12. H. A. Mohammed and I. A. M. Abdelrahman, "Detection and classification of malaria in thin blood slide images," in 2017 Int. Conf. Commun., Control, Comput. Electron. Eng. (ICCCCEE), 2017, pp. 1–5.
13. P. A. Pattanaik, M. Mittal, and M. Z. Khan, "Unsupervised deep learning CAD scheme for the detection of malaria in blood smear microscopic images," IEEE Access, vol. 8, pp. 94936–94946, 2020.
14. M. Turuk, R. Sreemathy, S. Kadiyala, S. Kotecha, and V. Kulkarni, "CNN-based deep learning approach for automatic malaria parasite detection," IAENG Int. J. Comput. Sci., vol. 49, no. 3, 2022.
15. V. M. Sibinraj, F. Gafoor, P. K. Anandhu, and V. S. Anoop, "Automated malaria cell image classification using convolutional neural networks," Authorea Preprints, 2024.